\documentclass[sigconf]{acmart}

\usepackage{graphicx}
\usepackage{multicol}
\usepackage{multirow}
\usepackage{epstopdf}
\usepackage{epsfig}
\usepackage{subfigure}
\usepackage[boxed,ruled]{algorithm2e}
\usepackage{setspace}
\usepackage{booktabs,bm} % For formal tables
\renewcommand\footnotetextcopyrightpermission[1]{}
\usepackage{url}
\usepackage{dirtytalk}
\usepackage{hyperref}
\usepackage{enumitem}
\acmYear{}
\acmPrice{}
\acmISBN{}
\acmDOI{}
\setcopyright{none}
\acmConference[AdvML'19: Workshop on Adversarial Learning Methods for Machine Learning and Data Mining at KDD]{AdvML'19: Workshop on Adversarial Learning Methods for Machine Learning and Data Mining at KDD}{August 5th, 2019}{Anchorage, Alaska, USA}

\title{Block Switching: A Stochastic Approach for Deep Learning Security}
\author{Xiao Wang$^{1*}$, Siyue Wang$^{2*}$,  Pin-Yu Chen$^{3}$, Xue Lin$^{2}$, and Peter Chin$^{1}$}
\affiliation{%
  \institution{1. Boston University \ 2. Northeastern University \ 3. IBM Research \ $^*$ Equal Contribution \\}
  \city{ \ kxw@bu.edu \quad wang.siy@husky.neu.edu \quad pin-yu.chen@ibm.com \quad xue.lin@northeastern.edu \quad spchin@cs.bu.edu }
}

\begin{abstract}
Recent study of adversarial attacks has revealed the vulnerability of modern deep learning models. That is, subtly crafted perturbations of the input can make a trained network with high accuracy produce arbitrary incorrect predictions, while maintain imperceptible to human vision system. In this paper, we introduce Block Switching (BS), a defense strategy against adversarial attacks based on stochasticity. BS replaces a block of model layers with multiple parallel channels, and the active channel is randomly assigned in the run time hence unpredictable to the adversary. We show empirically that BS leads to a more dispersed input gradient distribution and superior defense effectiveness compared with other stochastic defenses such as stochastic activation pruning (SAP). Compared to other defenses, BS is also characterized by the following features: (i) BS causes less test accuracy drop; (ii) BS is attack-independent and (iii) BS is compatible with other defenses and can be used jointly with others.
\footnotetext{This work is supported by the Air Force Research Laboratory FA8750-18-2-0058.}

\end{abstract}

\begin{document}

\maketitle

%%%%%%%%% Paper Content
\section{Introduction}
% Adversarial attacks.
% Optimization based attacks.
% Adversarial defenses.
% Defense by adding randomness.
% Our Methods, switching among sub models.
% Our results shows it reduces the attack succ rate from 100\% to ~20\%.
% Our contribution: come up with the switching methods. Explaining why it works better.
% Organizing of the paper.

Powered by rapid improvements of learning algorithms \cite{he2016deep,lecun2015lenet,krizhevsky2012imagenet,Zhao2019fault,zhao2018admm}, computing platforms \cite{abadi2016tensorflow,jia2014caffe}, and hardware implementations \cite{han2016eie,li2019rnn}, deep neural networks become the workhorse of more and more real world applications, many of which are security critical, such as self driving  cars \cite{bojarski2016end} and image recognition \cite{parkhi2015deep,he2016deep,krizhevsky2012imagenet,zhao2017aircraft,wang2018using}, where malfunctions of these deep learning models lead to serious loss.

However, the vulnerability of deep neural networks against adversarial attacks is discovered by Szegedy et al. \cite{szegedy2013intriguing}, who shows that in the context of classification, malicious perturbations can be crafted and added to the input, leading to arbitrary erroneous predictions of the target neural network. While the perturbations can be small in size and scale or even invisible to human eyes. 

% can fool a well-trained neural network, despite the high testing accuracy, to mis-classify the input with high confidence. While the perturbations can be small enough to be invisible to human eyes. 

% However, Szegedy et al. found that the reliability of the deep neural networks are shown to be threatened by adversarial attacks \cite{szegedy2013intriguing}. 
% % However, a potential threat to the liability of deep neural networks is discovered by Szegedy et al. \cite{szegedy2013intriguing}, who introduced the concept of adversarial attack for the first time in their work . 
% They showed that in the context of image classification, intentionally designed perturbations added to the input image, can fool a well-trained neural network, despite the high testing accuracy, to mis-classify the input with high confidence. While the perturbations can be small enough to be invisible to human eyes. 

% Although the concept of adversarial attack has been introduced in the contents of image classification, it threatens the reliability of deep neural networks in general. 

This phenomenon triggered wide interests of researchers, and a large number of attacking methods have been developed. Some typical attack methods include Fast Gradient Sign Method (FGSM) by Goodfellow et al.  \cite{Goodfellow2015explaining}, Jacobian-based Saliency Map Attack (JSMA) by Papernot et al.  \cite{papernot2016limitations}, and CW attack by Carlini and Wagner \cite{carlini2017towards}. 
% Although these attacking methods are different in terms of speed, attacking strength, and optimization objectives, they share the same ideology of generating adversarial examples. 
These attacks utilize gradients of a specific object function with respect to the input, and design perturbations accordingly in order to have a desired output of the network.
Among the attacks, CW attack is known to be the strongest and often used as a benchmark for evaluating model robustness.

% A common approach of all these methods is that they all require gradients backpropagated to the input, and use this gradient information to update the input to make it adversarial. 

In the meantime, a rich body of defending methods have been developed, attempting to improve model robustness in different aspects.
Popular directions include adversarial training \cite{madry2017towards}, detection \cite{grosse2017statistical,metzen2017detecting}, inputs rectifying \cite{das2017keeping,xie2017mitigating}, and stochastic defense \cite{s.2018stochastic, wang2018defensive, wang2018defending, wang2019protecting}. However, although these defenses alleviate the vulnerability of deep learning in some extent, they are either shown to be invalid against counter-measures of the adversary \cite{carlini2017adversarial} or require additional resources or sacrifices. A significant trade-off of these methods is between defense effectiveness and test accuracy, where a stronger defense is often achieved at the cost of worse performance on clean examples\cite{wang2019protecting}. 

% Adversarial training, although shown to be effective, is often referred as a "brute-force" method since it requires extremely larger amount of energy for training \cite{akhtar2018threat}. Detection methods are easily circumvented if the attacker count the detection network as part of target network \cite{carlini2017adversarial} and inputs rectifying methods often harms the testing accuracy as it corrupts the inputs structure \cite{xu2017feature}. On the other hand, network randomization methods defend against adversarial attacks by sampling model parameters from a certain distribution in the run time thus prevent the attacker from utilizing the vulnerability of a fixed network. These randomization methods are often easily implementable but effective in defending adversarial attacks.

% Stochastic Activation Pruning (SAP) \cite{dhillon2018stochastic} is a typical defense method with gradient obfuscating published recently. It builds a stochastic model by dropping neurons with certain probabilities. It is shown to be effective in defending FGSM. 

Motivated by designing defense method with less harm on test accuracy, in this article we introduce Block Switching (BS) as an effective stochastic defense strategy against adversarial attacks. BS involves assembling a switching block consisting of a number of parallel channels. Since the active channel in the run time is random, it prevents the adversary from exploiting the weakness of a fixed model structure. On the other hand, with proper training, the BS model is capable of adapting the switch of active channels, and maintains high accuracy on clean examples. As a result, BS achieves drastic model variation, and thus have strong resistance against adversary without noticeable drop in legitimate accuracy. The nature of BS also enables its usage jointly with other type of defenses such as adversarial training. 

% propose another randomization strategy for the sake of robustness against adversarial attacks which outperforms recently published randomization method such as SAP and defensive dropout. This strategy involves assembling a block switching consisting of a bunch of paralleled channels. %\textcolor{red}{(i.e., seperate bottom models)}
% In the inference phase, the input of the model is randomly assigned to one of the channels by the input switcher. While all channels lead to similar predictions for legitimate examples, it dramatically perturbs the gradient signal used by the attacker due to different weight parameters in different channels. 
% % connecting to a common upper model, and a switcher assigning the input to one of the channels randomly. We call the selected channel by the switcher \say{active channel}. Training process of block switching contains two rounds. In the first round, a bunch of sub-models (the effects of the number of sub-models are discussed later in Section 4.3) are trained which provide weights to initialize different channels. In the second round, the whole block switching is trained using the original training data that forces the commom upper model to learn to accommodate outputs from different channels. Usually the second round of training is much faster than the first round. In the inference phase, since the activate channel is ever-changing, it prevents the attacker from generating successful adversarial examples using the deterministic gradient signal during backpropagation. 

Our experimental results show that a BS model with 5 channels can reduce the fooling ratio (the percentage of generated adversarial examples that successfully fool the target model) of CW attack from 100\% to 21.0\% on MNIST dataset and to 22.2\% on CIFAR-10 dataset respectively with very minor testing accuracy loss on legitimate inputs. As comparison, another recent stochastic defense stochastic activation pruning (SAP) only reduces the fooling ratio to 32.1\% and 93.3\% given the same attack. The fooling ratio can be further deceased with more parallel channels. 

% In this paper, we make the following contributions:
% \begin{itemize}
%     \item{We introduce block switching and show its effectiveness in defending adversarial attacks.}
%     \item{We provide an analysis on defense of stochastic models in general and explain why block switching achieves better defending performance than other existing stochastic models.}
%     % the existing stochastic model Stochastic Activation Pruning (SAP)
  
% \end{itemize}

The rest of this article is organized in the following way: In Section 2, we introduce related works in both attacking and defending sides. The defense strategy and analysis are given in Section 3. Experimental results are given in Section 4. And Section 5 concludes this work. 

%==========================================================
% \vspace{-5mm}
\section{Adversarial Attack}
\noindent{\textbf{FGSM}.} Fast Gradient Sign Method (FGSM) \cite{Goodfellow2015explaining} utilizes the gradient of the loss function to determine the direction to modify the pixels. They are designed to be fast, rather than optimal. 

% They can be used for adversarial training by directly changing the loss function instead of explicitly injecting adversarial examples into the training data.

Specifically, Adversarial examples are generated as following:
\begin{equation}
x' = x - \epsilon \cdot \text{sign} (\nabla(loss_{F,t}(x)))
\label{FGSM}
\end{equation}
where $\epsilon$ is the magnitude of the added distortion, $t$ is the target label. Since it only performs a single step of gradient descent, it is a typical example of \say{one-shot} attack.

% \subsubsection{Jacobian-based Saliency Map Attack (JSMA)}
% JSMA \cite{papernot2016limitations} is an $L_0$ attack using a greedy algorithm that picks the most influential pixels by calculating Jacobian-based Saliency Map and modifies the pixels iteratively. The computational complexity of this attack method is very high.

\noindent{\textbf{CW}.} 
 Carlini \& Wagner (CW) attack \cite{carlini2017towards} generates adversarial examples by solving the following optimization problem:
\begin{equation}
\begin{array}{l}
\text{minimize} \quad D(\delta) + c\cdot f (x+\delta) \\
\text{subject to} \quad  x+\delta \in [0,1]^n \\
\end{array}
\label{cw object}
\end{equation}
where $c>0$ controls the relative importance between the distortion term $D$ and loss term $f$. The loss term $f$ takes the following form:
\begin{equation}
  f(x+\delta) = \text{max} (\text{ max}\{Z(x+\delta)_i : i \neq t\} - Z(x+\delta)_t, -\kappa)   
\end{equation}
where $\kappa$ controls the confidence in attacks.

\section{Method}

\subsection{Block Switching Implementation}
% (1) train individually
% (2) remove the top layer, add a common top layer
% (3) retrain the network.
% (4) Stochastic network

% A block switching is made with a number of pre-trained sub-models. There are two phases of training required in assembling a block switching.

Training a Block Switching model involves two phases. In the first phase, a number of sub-models with the same architecture are trained individually from random weights initialization. With the training process and data being the same, these models tend to have similar characteristics in terms of classification accuracy and robustness, yet different model parameters due to random initialization and stochasticity in the training process.

% %(we refer to as sub-models in order to avoid ambiguity with the overall block switching)  
% are trained individually in conventional manner. With the same training settings and data, they tend to have similar classification outputs while different weight parameters due to the random initialization and stochastic factors in training. 

\begin{figure}[htbp]
\centering
\vspace{-2mm}
\subfigure{
\centering 
\begin{minipage}[b]{0.4\textwidth}
\centering  
\includegraphics[width=1\textwidth]{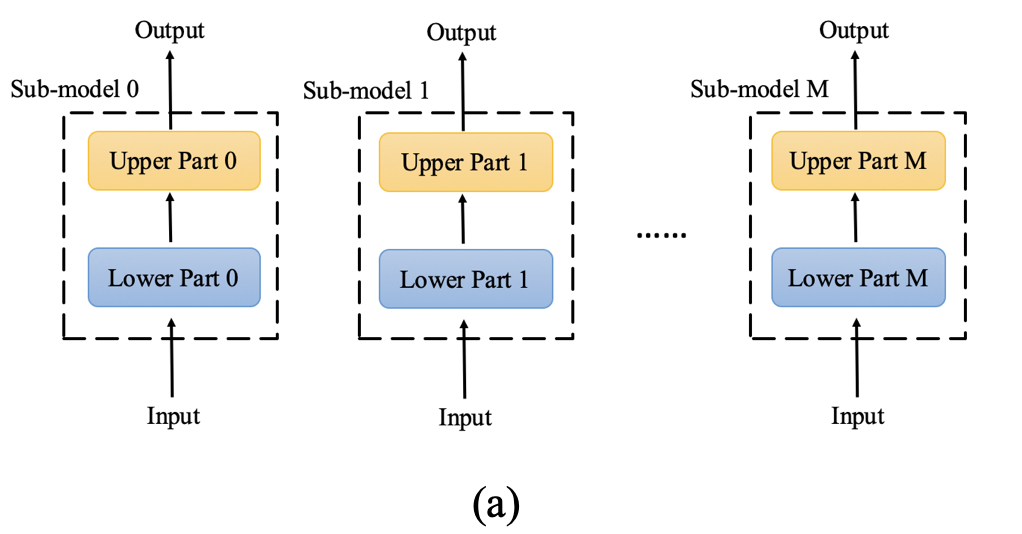} \\
\end{minipage}
}
\vspace{-4mm}
\subfigure{
\centering 
\begin{minipage}[b]{0.25\textwidth}
\centering  
\vspace{-3mm}
\includegraphics[width=1\textwidth]{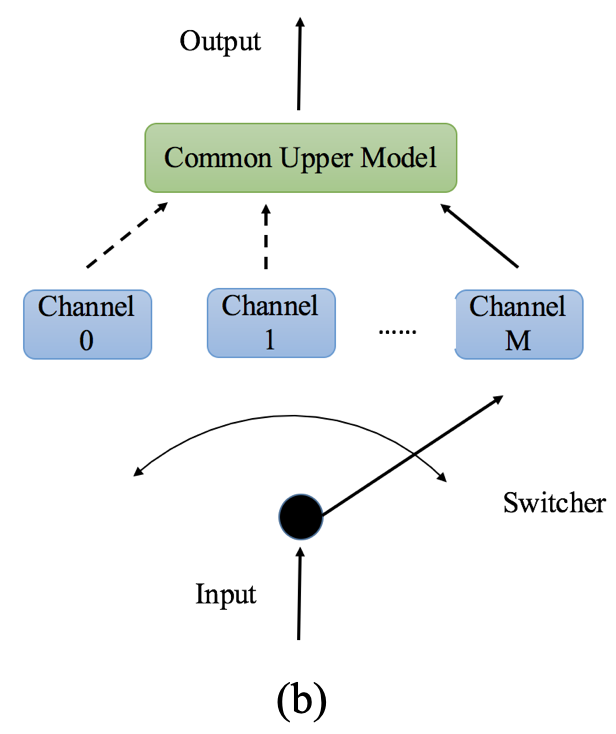} \\

\end{minipage}
\vspace{-5mm}
}
\caption{\textbf{The steps of assembling a block switching.} (a): Sub-models are trained individually. (b): The lower parts of sub-models are used to initialize parallel channels of block switching.}
\vspace{-6mm}
\label{Fig:MODEL_Switch} 
\end{figure}

After the first round of training, each sub-model is split into two parts. The lower parts are grouped together and form the parallel \textbf{channels} of the switching block, while the upper parts are discarded. The switching block is then connected to a randomly initialized common upper model as shown in Fig. \ref{Fig:MODEL_Switch}. In the run time, a random channel is selected to be \textbf{active} that processes the input while all other channel remains inactive, resulting in a stochastic model that has different behavior at different time.

% We then separate each trained sub-models into two parts. We call the part containing the input layer the lower part, and the remaining containing output layer the upper part. We discard the upper parts of all sub-models but place the lower parts parallelly and connect them to a randomly initialized common upper model that takes the same structure as the upper part of a sub-model. We refer to the lower parts of the sub-models as different \say{channels} of block switching. An input switcher is added to the model input gate which controlling assigning of the input image to one of the channels randomly in the run time. We call a channel \say{active} if it is selected by the switcher and inactive otherwise. An illustration of block switching structure is shown in Fig. \ref{Fig:MODEL_Switch}. 

% In general, we are taking the lower parts and replacing the upper parts with a common upper model and then retrain. This assembling process is shown in Fig. \ref{Fig:MODEL_Switch}.

% \begin{figure}[htbp]
% \centering  
% \begin{minipage}[c]{1\textwidth}
% \scalebox{0.45}{
% \centering                                  \includegraphics[width=1\textwidth]{SMP1.png}
% \centering                                  \includegraphics[width=1\textwidth]{SMP2.png}}
%  \end{minipage}
% \caption{\textbf{The steps of assembling a block switching.} (a): Sub-models are trained individually. (b): The lower parts of sub-models are used to initialize parallel channels of block switching.}
% \label{Fig:MODEL_Switch} 
% \end{figure}

The whole BS model is then trained for the second round on the same training dataset in order to regain classification accuracy. In this phase, the common upper model is forced to adapt inputs given by different channels so that a legitimate example can be correctly classified given whichever channel is active. Usually, this phase is much faster than the first round of training since the parallel channels are already trained. 
% In our experiments, block switching restores its testing classification accuracy to similar level as a regular model in 1 epoch of training on MNIST dataset and 5 epochs on CIFAR-10 dataset.

% The above process results in a stochastic model. Given the an input image, the output of the model is a random variable which can be denoted as:
% \begin{equation}
%     Y = \widetilde{F}(x)
% \end{equation}
% Here we follow the notations introduced in Section 2, but add a tilde on top of the model if it is stochastic.

\subsection{Defense Analysis}
% two power source:
% (1) the model changes 
% comparing to SAP, the model changes more 
% (2) Stochastic gradient
% comparing to SAP, the gradients vary more

Let $Y = \widetilde{F}(x)$ denoted the learned mapping of a stochastic model. Note that $\widetilde{F}$ is a stochastic function and now $Y$ is a random variable. The defending against adversarial attacks can be revealed in two aspects.
% There are two chains of reasoning to interpret the defense effectiveness of stochastic models.
\begin{itemize}[leftmargin=*]
    \item \textbf{Stochasticity of Inference}: Since $Y = \widetilde{F}(x)$ is a random variable, an adversarial example that fools an instance $F^1$ of the stochastic model $\widetilde{F}$ sampled at $t_1$ may not be able to $F^2$ sampled at $t_2$.
    \item \textbf{Stochasticity of Gradient} Due to the stochasticity of the network, the gradient of attacker's objective loss with respect to the input is also stochastic. That is, the gradient backpropagated to the input is just an instance sampled from the gradient distribution. And this instance may not represent the most promising gradient descent direction.
\end{itemize}

Note that these two aspects are actually correlated. From the attacker's point of view, the goal is to find $\arg \mathop {\max }\limits_x {\mathbb{E}[A(\widetilde{F}(x), T)]}$ where $A(\cdot)$ outputs 1 if the attack is successful and 0 otherwise, and $T$ is the target class. Therefore, the attacker is benefited from using stochastic gradients other than gradients from a fixed model instance, in order to generate adversarial examples that are robust to model variation. In another word, this means the adversary cannot benefit from simply disabling the variation of the stochastic model and craft perturbations using a fixed model instance.

% the stochastic property of the model cannot be circumvented by the attacker for better attacking strength and the benefits from using stochastic models are guaranteed to the defender.

% Nonetheless, stochastic gradients are not good for nothing. A small amount of gradient perturbation widens the searching area during gradient descent by occasionally by heading to a less \say{correct} direction while a large perturbation may hurt the descending. This trade-off is usually referred to as exploration-exploitation trade-off.

The above analysis holds for any stochastic model but the question is what makes a good randomizatin strategy against adversarial attacks? Intuitively, a good randomization strategy should cause the input gradients to have wider distributions. In an extreme case, if the gradient direction is uniformly distributed, performing gradient descent is no better than random walking, which means the attacker cannot take any advantage from the target model. 

% The above analysis can be applied to all stochastic models while the stochasticity can be introduced by different manners, e.g. by adding Gaussian noise to layer outputs as in  \cite{nguyen2017learning} or by randomly masking activations as in SAP \cite{s.2018stochastic}. But what are good stochastic strategies against adversarial attacks? Intuitively, a good stochastic defending scheme should make the model to have large variations so that the probability distribution of the gradients used when generating adversarial examples become wider. Let us think about an extreme situation. If the direction of gradient is uniformly distributed, then the gradient descent step will perform the same action as a random walk, which means the attacking algorithm cannot benefit from knowing the target model. 

\begin{figure}
    \centering
    \subfigure{
\centering
\vspace{-4mm}
\begin{minipage}[b]{0.45\textwidth}
\centering  
\includegraphics[width=1\textwidth]{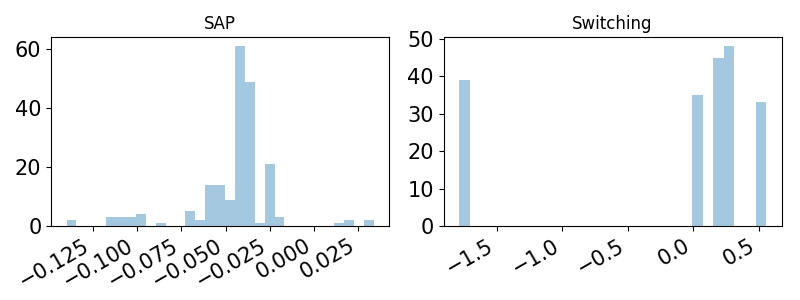} \\
\end{minipage}
}
    \subfigure{
\centering 
\begin{minipage}[b]{0.45\textwidth}
\centering
\vspace{-4mm}
\includegraphics[width=1\textwidth]{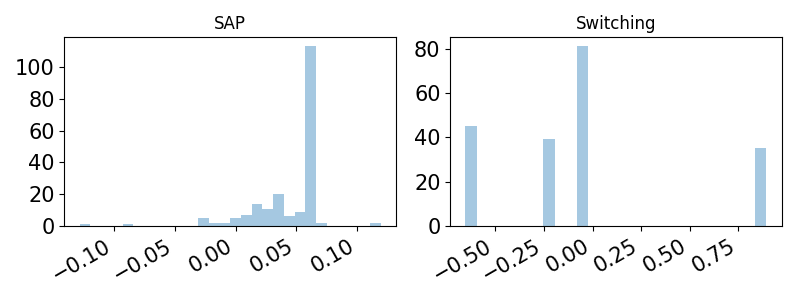} \\
\end{minipage}
}
    \subfigure{
    \vspace{-6mm}
\centering 
\begin{minipage}[b]{0.45\textwidth}
\centering  
\vspace{-4mm}
\includegraphics[width=1\textwidth]{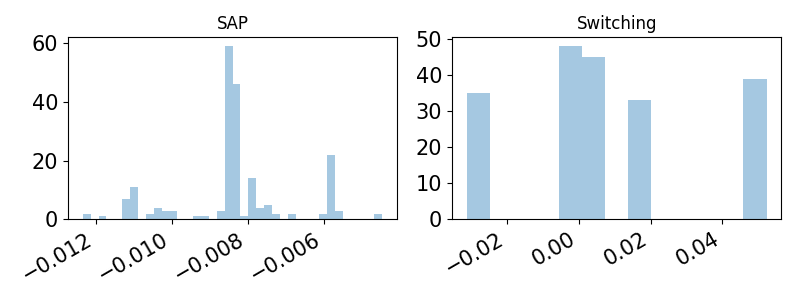} \\
\end{minipage}
}
% \vspace{-4mm}

    \caption{We use three images (a-c): Gradient distributions of CW attack on a SAP model. (d-f): Corresponding gradient distributions on a block switching. Distributions in the same column belong to the same input dimension. Each distribution is sampled for 100 times.}
    \label{SAP}
     \vspace{-5mm}
\end{figure}

Knowing this, we explain why block switching performs better than existing stochastic strategies such as SAP. In Fig. \ref{SAP} we visualize gradient distributions under CW attacks to a SAP model and a BS model respectively. We observe that the gradient (of the attacker's object function w.r.t the input) distribution of the SAP model is unimodal and concentrated While the gradient of BS has a multimodal distribution in a wider range. This distribution indicates that it is harder to attack BS than SAP which is verified by our experiment results in Section 4.
% obfuscates the gradient signal and hardens the process of generating adversarial examples. 

% bell shape, which stems from the nature of SAP, as shown in Eqn. \ref{sap}, neurons with smaller activation outputs are more likely to be dropped, which means small gradient variation is more likely to happen than large variation. On the other hand, the gradient distribution on block switching is wider and has a clear multimodal pattern. This is because block switching varies greatly when the active channel alternates.

% \begin{figure}
%     \centering
%     \includegraphics{}
%     \caption{Caption}
%     \label{fig:my_label}
% \end{figure}

% \begin{figure*}[htbp]
% \centering                    
% \begin{minipage}[c]{1\textwidth}
% \centering                                  \includegraphics[width=1\textwidth]{SAP.png}
% \centering                                  \includegraphics[width=1\textwidth]{SAP_SM.png}
%  \end{minipage}
% \caption{ We use three images (a-c): Gradient distributions of C\&W attack on a SAP model. (d-f): Corresponding gradient distributions on a block switching. Distributions in the same column belong to the same input dimension. Each distribution is sampled for 100 times.}
% \label{SAP} 
% \end{figure*}

Usually dramatic variations of the stochastic model tend to harm classification accuracy on clean inputs. That is why in SAP, smaller activation outputs have more chance to be dropped. The reason that Block Switching maintain high test accuracy despite drastic model change is that, since each channel connected to the common upper model is able function independently. As long as the common upper model can learn to adapt different knowledge representations given by different channels, the stochastic model will not suffer from significant test accuracy loss. 

% However, this limitation is largely relaxed during the second round of training of block switchings.

An interesting question that readers may ask is: why stochasticity of the model does not impede the second round of training? The fact is that although the gradients with respect to the input are random variables, the gradients with respect to model parameters are not. Since gradients of the inactive channel are just zeros, only weights parameters in the activate channel will be updated in each training step. Therefore, although the weights to be updated alternates, the gradients with respect to model parameters are deterministic at any time.
% This is one of clever designs of block switching.
%------------------------------------------------------------------------- 

\section{Experiments}
In this section, we compare the defense effectiveness of regular, SAP and BS models against FGSM \cite{Goodfellow2015explaining} and CW \cite{akhtar2018threat} attacks on MNIST \cite{lecun1998mnist} and CIFAR-10 \cite{krizhevsky2009learning} datasets. 
FGSM is a typical \say{one-shot} method which performs only one gradient descent step and CW attack is known to be the strongest attack method so far \cite{akhtar2018threat}. 

Both of these two datasets contain separated training and testing sets. In our experiments, the training sets are used to train the defending models and the testing sets are used to evaluate classification performance and generate adversarial examples. 

This section is organized in the following way: Details about the defending models, including the models' architectures and training methods, are given in Section 4.1. Defending records against FGSM and CW attacks are shown in Section 4.2. Study on how the number of channels in the block switching influences its the defending effectiveness and classification accuracy is provided in Section 4.3. 

\subsection{Model Details}
\subsubsection{Regular Models}
We use two standard Convolutional Neural Networks (CNNs) architectures for MNIST and CIFAR-10 datasets respectively, as they serve as baseline models repeatedly in previous works \cite{papernot2016distillation}. Both of these two CNNs have 4 convolutional layers, 2 pooling layers and 2 fully-connected layers but the kernel size of convolution filters and layer width are different.

% The architectures of these two models are shown in Fig. \ref{Fig:MODEL_Architecture}.

%==========================================================
% \begin{figure}[htbp]
% \centering                    
% \begin{minipage}[c]{1\textwidth}
% % \subfigure[MNIST Model Architectuer]{
% \centering                                  \includegraphics[width=1\textwidth]{MNIST_architecture.png}
% % \subfigure[CIFAR-10 Model Architectuer]{
% \centering                                  \includegraphics[width=1\textwidth]{Cifar_architecture.png}
% \end{minipage}
% \caption{ Above: MNIST model architecture. Below: CIFAR-10 model architecture. }
% \label{Fig:MODEL_Architecture} 
% \end{figure}
%==========================================================

Both models are trained using stochastic gradient descent with the mini batch size of 128. Dropout \cite{srivastava2014dropout} is used as regularization during training.

\subsubsection{SAP}
SAP can be applied post-hoc to a pre-trained model \cite{dhillon2018stochastic}. Therefore, in order to make the experimental results more comparable, we use the same trained weights for SAP model as of the regular model. Stochastic activation pruning is added between the first and second fully-connected layers.
%==========================================================
%==========================================================
\subsubsection{Block Switching}
The switching block in this experiment consists of 5 channels. During the first round of training, 5 regular models are trained as described above. Each regular model is split into a lower part, containing all convolutional layers and the first fully-connected layer, and a upper part, containing the second fully-connected layer. 
The lower parts of regular model are kept, providing parallel channels of block switching while the upper parts are discarded. A upper model, which is the same as the upper part of regular models except that its weights are randomly initialized, is added on top of all channels. The whole block switching is then trained on original training set for the second time. We found that the second round of training is much faster than the first round. On MNIST dataset block switching is retrained for 1 epoch and on CIFAR-10 dataset 5 epochs.

\begin{table}[ht]
\begin{center}
\vspace{-2mm}
\caption{Testing Accuracy of different models on MNIST and CIFAR-10 datasets.}
\scalebox{0.9}{
\begin{tabular}{|l|c|c|}
\hline
Model & Test Acc. on MNIST & Test Acc. on CIFAR \\
\hline\hline
Regular & 99.04\% & 78.31 \% \\
SAP & 99.02\% & 78.28 \% \\
Sub-models Avg. & 99.02\% & 78.97\%\\
Switching & 98.95\% & 78.73\%\\
\hline
\end{tabular}}
\end{center}
% \caption{Testing Accuracy of different models on MNIST and CIFAR-10 datasets.}
\label{AAA}
\end{table}

The test classification accuracy of all models is summarized in Table 1. The direct comparisons are between the regular model and the SAP model, since they share the same weights; and the average of sub-models used to construct block switching and block switching itself. We can conclude that both SAP and block switching are excellent in maintaining testing accuracy. 

% An interesting fact we observe is that on Cifar-10 dataset the testing accuracy of the swithcing model is even higher than the average of all submodels. An explanation of this is block switching retraining with alternating channels may play a role similar to data augmentation and hence improve the model's robustness.

\subsection{Defense against Adversarial Attacks}
We use the fooling ratio, which is the percentage of adversarial examples generated by a attack method that successfully fools a neural network model to predict the target label, to evaluate the defense effectiveness of the target model. The lower the fooling ratio is, the stronger the model is in defending adversarial attacks.

We also record the average $L_2$ norm of the generated adversarial examples from legitimate input images, since it is only fair to compare two attacks at similar distortion levels. For attacks like CW attack that uses a leveraged object function between distortion and misclassification, a large distortion also indicates that it is hard for the attacking algorithm to find an adversarial example in a small region. 

% For the sake of reproducibility of our experiments, we report the hyper-parameter settings for each attacks we use. 

\subsubsection{Experiments on MNIST Dataset}

For the sake of reproducibility of our experiments, we report the hyper-parameter settings we use for FGSM and CW attacks. FGSM has one hyper-parameter, the attacking strength $\epsilon$ as shown in equation \ref{FGSM}. 
% We do FGSM experiments using two attack strength settings, $\epsilon=0.1$ and $\epsilon = 0.25$ respectively. 
When using $\epsilon=0.1$, the $L_2$ norm of adversarial examples roughly matches CW, but the fooling ratio is way too small. Thus we also test the case when $\epsilon = 0.25$ in order to provide a more meaningful comparison, although the $L_2$ norm is significantly larger. For CW attack, gradient descent is performed for 100 iterations with step size of 0.1. The number of binary searching iterations for $c$ in \ref{cw object} is set to 10.
% For CW\_SAP, we find that the clipping on the pixel value difference between adversarial example and original image significantly decrease the fooling ratio. This is because the successfully adversarial examples on MNIST usually have some pixels significantly changed \ref{} due to the nature the images in MNIST dataset, and clipping the pixel value difference significantly limits the attack. Since CW\_SAP was original designed on Cifar-10 but not MNIST, we omit the results of CW\_SAP on the MNIST dataset here.  

We use FGSM and CW attacks to generate adversarial examples targeting the regular model, the SAP model and block switching respectively. Experimental results are shown in Table \ref{defend MNIST}. 

\begin{table}[ht]
\centering
\caption{Fooling ratio (FR) and distortion of FGSM and CW attacks with different target models on MNIST dataset.}
\vspace{-1mm}
\label{Table: ASR_MNIST}
\scalebox{0.95}{
\begin{tabular}{|l|c|c|c|c|c|c|}
\hline
  \multirow{2}{*}{Attack}& \multicolumn{2}{|c|}{Regular} & \multicolumn{2}{|c|}{SAP} & \multicolumn{2}{|c|}{Switching} \\
\cline{2-7}
& FR & L2 & FR & L2 & FR & L2 \\
\hline
\hline
FGSM $\epsilon=0.1$ & 3.9\% & 2.73 & 3.7\% & 2.73 & 1.6\% & 2.73\\
\hline
FGSM $\epsilon=0.25$ & 34.0\% &6.84 & 32.8\% & 6.84 & 20.3\% & 6.84\\
\hline
CW & 100.0\% & 2.28 & 32.1\% & 2.28 & 21.0\% & 2.37\\
\hline
\end{tabular}} 
\label{defend MNIST}
\end{table}

Although the SAP model demonstrates its extra robustness against both FGSM and CW than the regular model, block switching is apparently superior and deceases the fooling ratio further.

\subsubsection{Experiments on CIFAR-10 Dataset}

We use $\epsilon=0.01$ for FGSM in this experiment in order to have adversarial examples with similar distortion level comparing to examples generated by CW attack. The hyper-parameter setting for CW attack is the same as above.

Experimental results on CIFAR-10 datasets are shown in Table 3. And block switching significantly decreases fooling ratio of FGSM and CW to 8.1\% and 22.2\% respectively while the SAP model only shows minor advantages over the regular model.

\begin{table}[ht]
\caption{Fooling ratio (FR) and distortion of FGSM and C\&W attacks with different target models on CIFAR-10 dataset.}
\vspace{-2mm}
\centering
\scalebox{0.95}{
\begin{tabular}{|l|c|c|c|c|c|c|}
\hline
%   \multirow{2}{3em}{Attack}&
  \multirow{2}{*}{Attack}& \multicolumn{2}{|c|}{Regular} & \multicolumn{2}{|c|}{SAP} & \multicolumn{2}{|c|}{Switching} \\
\cline{2-7}
& FR & L2 & FR & L2 & FR & L2 \\
\hline
\hline

FGSM $\epsilon=0.01$ & 25.0\% & 0.55 & 24.8\% & 0.55 & 8.1\% & 0.55\\
\hline
CW & 100.0\% & 0.54 & 93.3\% & 0.52 & 22.2\% & 0.69\\
\hline
\end{tabular}
}
% \caption{Fooling ratio (FR) and distortion of FGSM and C\&W attacks with different target models on CIFAR-10 dataset.}
\label{defend CIFAR}
\end{table}

\subsection{the Effect of Channel Number}

To provide an analysis on how the number of channels in a block switching affect its defense effectiveness as well as testing accuracy, we run CW attack on BS models with different number of channels ranging from 1 (which is a regular model) to 9.

\begin{figure}[ht]
    \centering
    \vspace{-1mm}
    \begin{minipage}[c]{1\textwidth}
    \scalebox{0.43}{
    \includegraphics[width=1\textwidth]{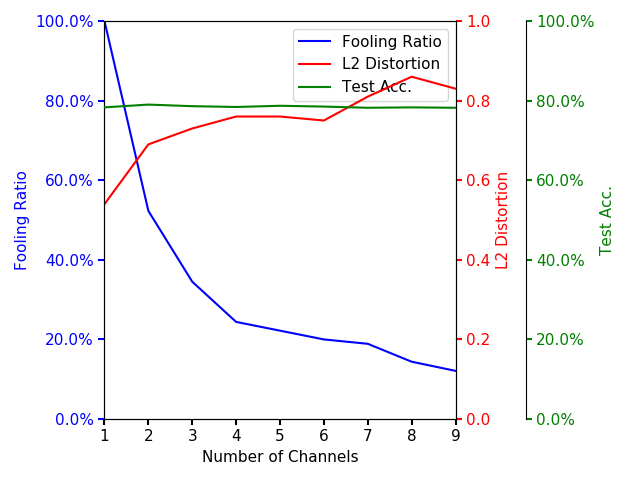}}
    \end{minipage}
    \vspace{-3mm}
    \caption{\textbf{Quantifying the impact of channel numbers:} we plot defending effectiveness in terms of fooling ratio and $L_2$ distortion, and testing classification accuracy of block switchings with 1 channel to 9 channels.}
    \label{channels}
    \vspace{-2mm}
\end{figure}

In Fig. \ref{channels} we plot the fooling ratio, distortion and test accuracy over different channel numbers: in general, the defense becomes stronger with more channels of block switching and the fooling ratio is lowest, 12.1\%, when using 9 channels. The fooling ratio drops rapidly from 1 channel to 4 channels while the drop of fooling ratio decelerates after 5 channels, which indicates the effectiveness provided by switching channels starts to saturate. The increasing of distortion of adversarial examples also indicates that BS with more channels are stronger when defending adversarial attacks. The trend of testing accuracy, on the other hand, is almost flat with a very slight descent from 78.31\% to 78.17\%. This indicates that BS is very effective in defending adversarial attacks with very minor classification accuracy loss.

%==========================================================%==========================================================%==========================================================

%==========================================================

% \subsection{Figures and Tables}
% \paragraph{Positioning Figures and Tables} Place figures and tables at the top and 
% bottom of columns. Avoid placing them in the middle of columns. Large 
% figures and tables may span across both columns. Figure captions should be 
% below the figures; table heads should appear above the tables. Insert 
% figures and tables after they are cited in the text. Use the abbreviation 
% ``Fig.~\ref{fig}'', even at the beginning of a sentence.

%==========================================================

\section{Conclusions}

In this paper, we investigate block switching as a defense against adversarial perturbations. We provide analysis on how the switching scheme defends adversarial attacks as well as empirical results showing that a block switching model can decease the fooling ratio of CW attack from 100\% to 12.1\% . We also illustrate that stronger defense can be achieved by using more channels at the cost of slight classification accuracy drop.

Block switching is easy to implement which does not require additional training data nor information about potential adversary. Also, it has no extra computational complexity than a regular model in the inference phase since only one channel is used at a time. In practice, the parallel channels can be stored distributedly with periodical updating, which can provide extra protection of the model that prevents important model information leak.

More importantly, BS demonstrates that it is possible to enhance model variation yet maintain test accuracy at the same time. And we hope this paper can inspire more works toward this direction.

%==========================================================
\bibliographystyle{ACM-Reference-Format}
\bibliography{KDD}

\end{document}